\documentclass[conference]{IEEEtran}
\IEEEoverridecommandlockouts
\usepackage{cite}
\usepackage{amsmath,amssymb,amsfonts}
\usepackage{algorithmic}
\usepackage{graphicx}
\usepackage{textcomp}
\usepackage{xcolor}

\usepackage{subcaption}
 \usepackage{dblfloatfix}

\usepackage{bigstrut}
\usepackage{multirow}
\usepackage{color}
\usepackage{caption}
\usepackage{fancyhdr}
\usepackage{listings} 
\usepackage{changepage}
\usepackage{pgfplots}
\usepackage{balance}
\usepackage{enumitem}
\usepackage{breqn}
\usepackage[pagebackref=true,breaklinks=true,letterpaper=true,colorlinks,bookmarks=false]{hyperref}
 
\begin{document}

\title{Vulnerability of Face Morphing Attacks: A Case Study on Lookalike and Identical Twins}

\author{Raghavendra Ramachandra\textsuperscript{1} \quad \quad Sushma Venkatesh\textsuperscript{2} \quad \quad   Gaurav Jaswal \textsuperscript{3} \quad \quad   Guoqiang Li \textsuperscript{4}\\
\textsuperscript{1}Norwegian University of Science and Technology (NTNU), Norway.
\textsuperscript{2}AiBA AS, Norway. \\
\textsuperscript{3}Indian Institute of Technology (IIT), Mandi.
\textsuperscript{4}MOBAI AS, Norway.\\
}
\IEEEoverridecommandlockouts
\IEEEpubid{\makebox[\columnwidth]{Paper Accepted in IWBF 2023 \hfill} \hspace{\columnsep}\makebox[\columnwidth]{ }}
\maketitle
\IEEEpubidadjcol
\begin{abstract}
Face morphing attacks have emerged as a potential threat, particularly in automatic border control scenarios. Morphing attacks permit more than one individual to use travel documents that can be used to cross borders using automatic border control gates. The potential for morphing attacks depends on the selection of data subjects (accomplice and malicious actors). This work investigates lookalike and identical twins as the source of face morphing generation. We present a systematic study on benchmarking the vulnerability of Face Recognition Systems (FRS) to lookalike and identical twin morphing images. Therefore, we constructed new face morphing datasets using 16 pairs of identical twin and lookalike data subjects.  Morphing images from  lookalike and identical twins are generated using a landmark-based method. Extensive experiments are carried out to benchmark the attack potential of lookalike and identical twins. Furthermore, experiments are designed to provide insights into the impact of vulnerability with normal face morphing compared with lookalike and identical twin face morphing.   
 
\end{abstract}

\begin{figure}[t!]
\begin{center}
\includegraphics[width=1.0\linewidth]{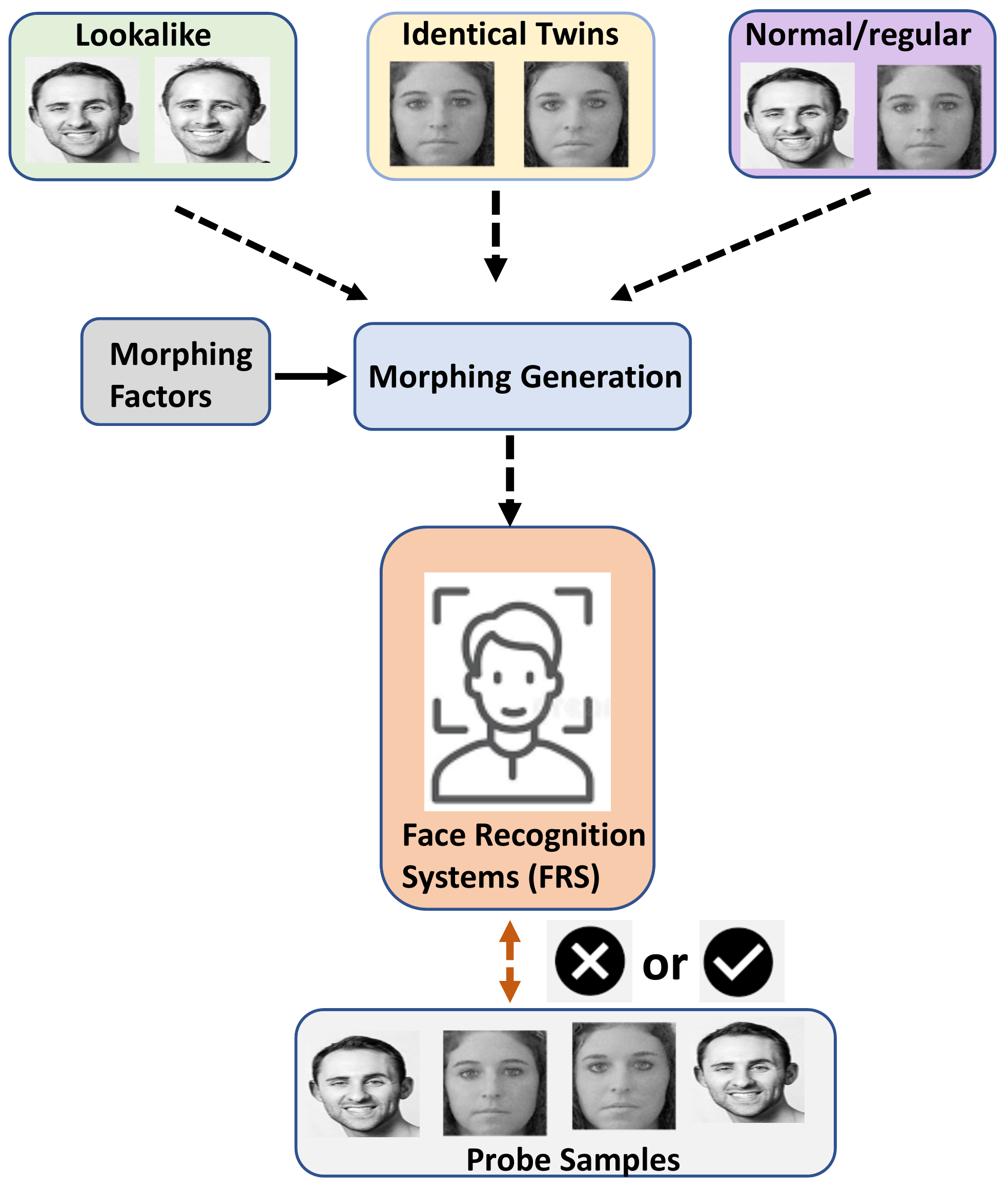}
\end{center}
   \caption{Illustration of the influence of lookalike and identical twins morphing on face recognition system}
\label{fig:Intro}
\end{figure}
\section{Introduction}
\label{sec:intro}
Biometric   person verification systems that use either physical or behavioral characteristics have been extensively deployed in  various applications, including border control. Facial biometrics are the primary identifiers in electronic passports (e-passports) that can enable automatic border control applications. The popularity of face biometrics in high-security applications can be attributed to user convenience, nonintrusive capture, and acceptable verification performance. However, face biometrics are highly vulnerable to presentation attacks in which attack instruments are generated using low-cost materials \cite{Raghavendra-FacePAD-Survey-2017, yu2022deep}. Morphing attacks have demonstrated high vulnerability among the different types of attacks, especially in passport issuance and automatic border control scenarios. 

Morphing is the process of seamless blending of two or more images, such that the resulting image shows visual similarities corresponding to the source images used for morphing. Face-morphing techniques blend two or more face images to generate a single-face image. Earlier studies have demonstrated that face morphing images indicate the vulnerability of the commercial Face Recognition System (FRS) \cite{Ferrara-TheMagicPassport-IJCB-2014}, deep learning-based FRS \cite{zhang-MIPGAN-TBIOM-2021}, and human observers \cite{godage2022analyzing}. Thus, the detection of morphing attacks on FRS has gained momentum, resulting in several techniques based on a single image and differential image \cite{Venkatesh-MADSurvey-IEEETTS-2021}. Even though the vulnerability of the FRS is well evaluated on normal (or regular) faces, it is an under-studied problem with lookalike and identical twin data subjects.    

The twin population across the globe has experienced a significant rise, equivalent to approximately 1.4 million children a year. This translates to one among the 45 who will be born as twins. Identical twin face recognition is a challenging problem, as FRS typically fails to distinguish between twins. In addition, lookalike (or doppelganger) face recognition is still a  challenging problem for FRS because it fails to differentiate owing to highly similar facial features. A study presented in \cite{lucas2015human, LookAlike1} indicated that one in 135 people could find a single identical lookalike. It was demonstrated in  \cite{joshi2022look} that three different FRS have limitations in verifying look-alike pairs. Twins and lookalike face recognition have been extensively studied in the biometric community. An early benchmark study \cite{phillips2011distinguishing} on identical twins outlined that the identical twin impostor distribution is more similar to the genuine distribution than the general impostor distribution. Three different FRS were evaluated under six different experimental conditions, indicating the challenge of reliable twin-face verification with varying image conditions. An extensive survey on identical twin face recognition was presented in \cite{bowyer2016biometric} and discussed techniques developed to improve the verification performance of identical twins. Recent approaches \cite{9024704, afaneh2017recognition} based on deep learning and Siamese networks have reported a marginal improvement in the verification performance of twin-face recognition.   

\begin{figure*}[htp]
\minipage{0.47\textwidth}
  \includegraphics[width=\linewidth]{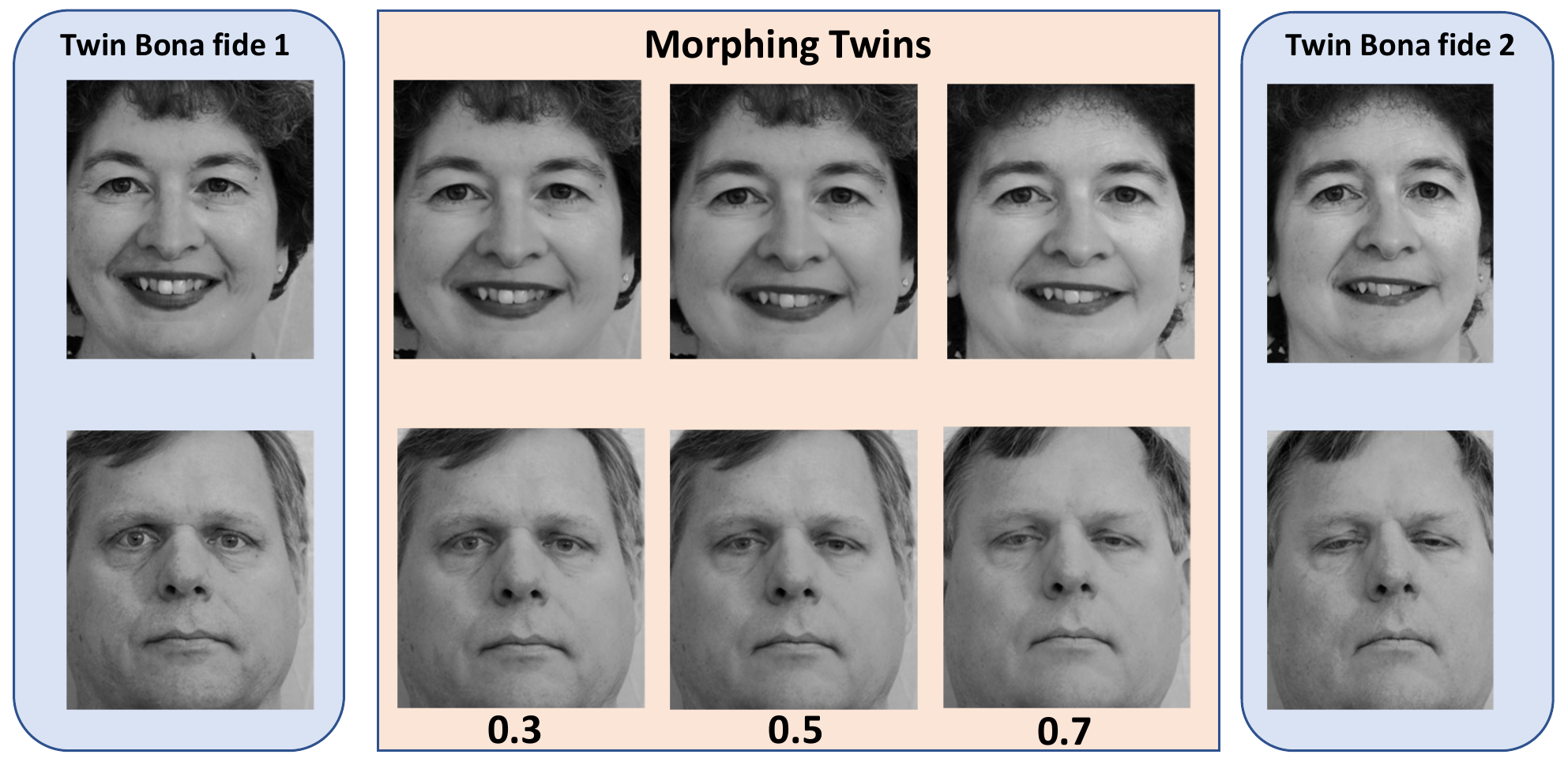}
  \caption{Identical Twins morphing examples}\label{fig:DBTwins}
\endminipage
\hfill
\minipage{0.47\textwidth}
  \includegraphics[width=\linewidth]{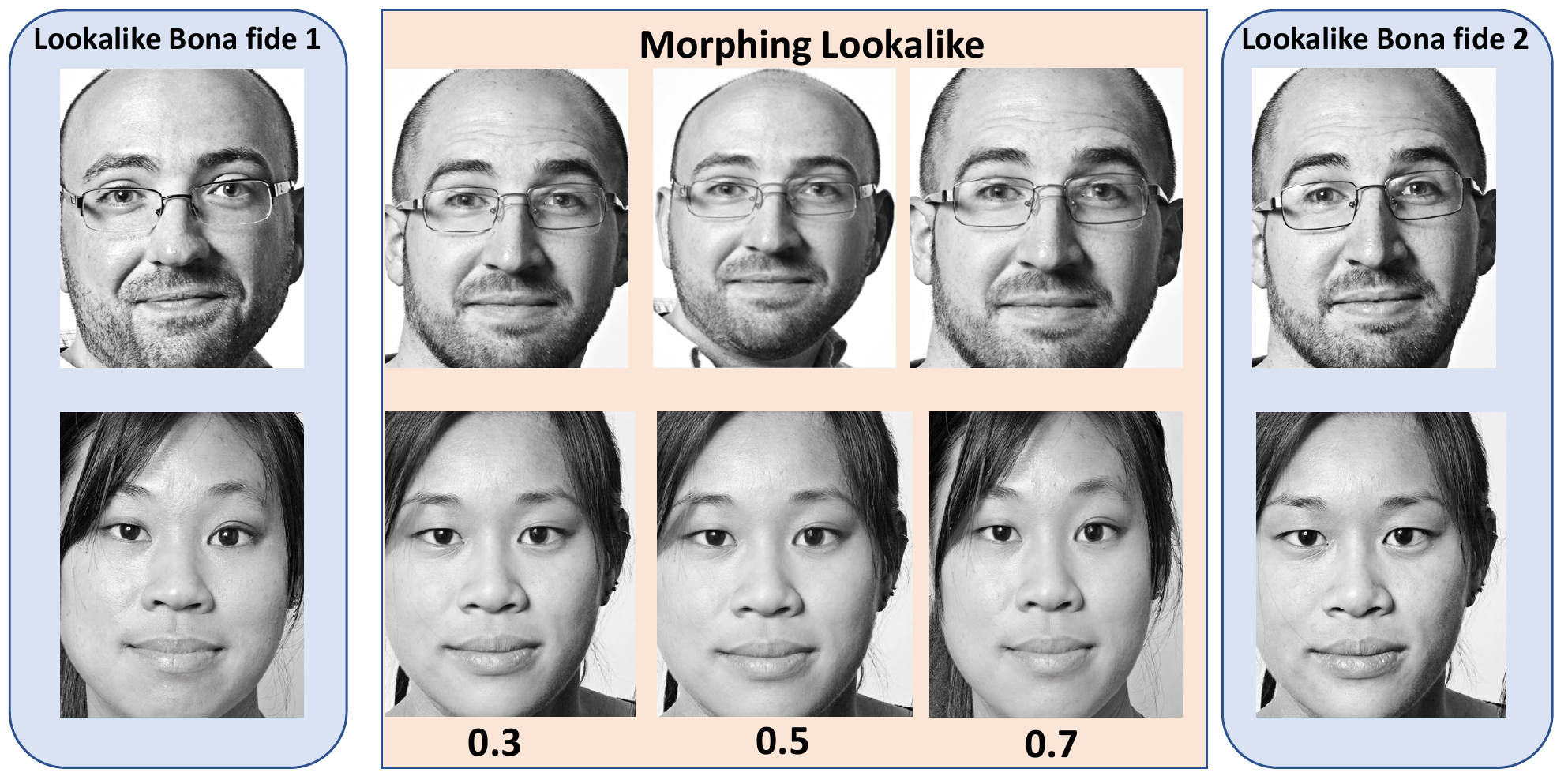}
  \caption{Lookalike morphing examples}\label{fig:awesome_image2}
\endminipage
\end{figure*}

Look-like face recognition has been well studied in the biometric literature. Early work \cite{lamba2011face} on lookalike face recognition showed a higher number of false matches. Extensive experiments are presented with ten different FRS and a new method based on the facial region to improve the face verification performance for lookalikes. Since then, several approaches have been proposed \cite{ moeini2017open, smirnov2017doppelganger, rustam2019application, swearingen2021lookalike} to enhance the performance of face recognition systems. However, it is worth noting that the datasets used in the literature are curated from the Web and thus have various image qualities. Recently, in \cite{joshi2022look}, a high-quality image database of lookalike data subjects with similar genetics was used to present the vulnerability of the FRS. Since the lookalike data subjects share similar genetics perceptually, they indicate a strong resemblance to each other.  

\subsection{Motivation and Contributions}

Identical twins and lookalikes have covered a reasonable population across the globe, and morphing attacks are highly vulnerable, especially in passport issuance and border control scenarios. The success of morphing attacks is highly reliable if an attacker can find a lookalike accomplice. Therefore, in this study, we are motivated to provide insight into the vulnerability of the FRS for both lookalike and identical twins. Figure \ref{fig:Intro} illustrates examples of lookalike and identical twins and their impacts on the vulnerability of the FRS. To the best of our knowledge, this is the first study to present insights into the vulnerability of FRS to the morphing of lookalike twins and identical twins. In particular, we introduce the following critical questions:
 \begin{itemize}[leftmargin=*,noitemsep, topsep=0pt,parsep=0pt,partopsep=0pt]
\item Does the morphing of lookalike, and identical twins indicate higher vulnerability of FRS compared to the normal (or regular) face?
    \item Does the morphing of lookalike data subjects indicate higher vulnerability of FRS  than identical twins?
    \item Does the morphing factor influence the vulnerability of FRS to lookalikes than identical twins?
    \item Does the Commercial-Off-The-Shelf (COTS) FRS indicate a higher vulnerability than deep learning FRS (Arcface)?
\end{itemize}
In the course of answering the above research questions, the following are the main contributions of this work:
\begin{itemize}[leftmargin=*,noitemsep, topsep=0pt,parsep=0pt,partopsep=0pt]
      \item  First work addressing the lookalikes and identical twins morphing attacks vulnerability on FRS.
   \item  Vulnerability analysis is presented using two different FRS, including COTS \cite{cognitec-FRS-SDK} \footnote{Disclaimer: These results were produced in experiments conducted by us and should; therefore, the outcome does not necessarily constitute the best the algorithm can do.} and deep learning (ArcFace \cite{Deng-ArcFace-IEEE-CVPR-2019}).
   \item  New morphing dataset corresponding to lookalikes and identical twins.
   \item Extensive experiments are presented to benchmark the vulnerability of lookalike and identical twins.  
\end{itemize}

The rest of the paper is organized as follows: Section  \ref{sec:db} discusses the details of the newly constructed morphing datasets using lookalikes and identical twins, Section \ref{sec:exp} presents a qualitative and quantitative analysis of the vulnerability analysis, and Section \ref{sec:conc} concludes the paper.  

\section{lookalike and identical twins morphing dataset}
\label{sec:db}
 
This section discusses the newly constructed face morphing dataset corresponding to identical twin and lookalike datasets. The lookalike face database employed in this work is based on a publicly available dataset \cite{LookAlikeDB, joshi2022look}. We mainly employ this database compared to other similar datasets because (a) image quality: images are captured under constrained conditions with uniform lighting and a professional photographer. However, the other existing datasets are harvested from a web source that has no control over the image quality (compression artifacts, capture with different cameras, and uncontrolled environmental conditions). (b) Genetically similar: The lookalikes employed in this work are proven to have similar genetics, and thus, they exhibit high likeness. However, other existing datasets have yet to prove to have similar genetics and, thus, do not necessarily exhibit high likeness. (c) Natural capture: The data subjects were captured naturally without extreme makeup. However, the data subjects in other similar datasets may have makeup, or images might have been processed to increase their appearance. The lookalike dataset employed in this study had 16 lookalike pairs that were used to generate the morphing image. In this study, we employed a landmark-based face morphing tool \cite{Ferrara-TextureBlendingAndShapeWarpingInFaceMorphing-IEEE-BIOSIG-2019} by considering its ability to generate high-quality morphing images, resulting in high vulnerability across different FRS \cite{zhang-MIPGAN-TBIOM-2021}. Morphing images were generated with three different morphing factors: 0.3, 0.5, and 0.7. Non-twin morphing was also generated to provide a comprehensive comparison.

The identical twin morphing dataset was generated based on a publicly available dataset from the University of Notre Dame Twins database \cite{phillips2011distinguishing}. To perform an effective comparative analysis with lookalike faces, we selected 16 identical twin pairs, particularly in the controlled scenario simulating the real-life scenario of face morphing. We then perform the morphing using a landmarks-based method that we used with a lookalike to generate the morphing images with three different morphing factors (0.3, 0.5, 0.7). In addition to  identical face morphing, we performed non-identical face morphing from the same dataset  to present a comprehensive comparison. Table \ref{tab:DBTable} shows the statistics of the newly constructed dataset, and Figure \ref{fig:DBTwins} and \ref{fig:awesome_image2} show an example of the newly generated morphing dataset corresponding to lookalike and identical twins.

\begin{table}[htbp]
  \centering
  \caption{Statistics of the newly constructed database}
  \resizebox{0.99\linewidth}{!}{
    \begin{tabular}{|p{7.285em}|c|c|c|c|}
    \hline
    \multirow{2}[4]{*}{Data Type} & \multicolumn{1}{c|}{\multirow{2}[4]{*}{No. of pairs}} & \multicolumn{1}{c|}{\multirow{2}[4]{*}{No. of Bona fide samples}} & \multicolumn{2}{p{8.43em}|}{No. of morphed samples} \bigstrut\\
\cline{4-5}    \multicolumn{1}{|c|}{} &       &       & \multicolumn{1}{p{4.215em}|}{Identical} & \multicolumn{1}{p{4.215em}|}{Non-Identical} \bigstrut\\
    \hline
    Lookalikes & 16    & 32   & 96    & 516 \bigstrut\\
    \hline
    Identical twins & 16  & 32    & 96    & 516 \bigstrut\\
    \hline
    \end{tabular}%
    }
  \label{tab:DBTable}%
\end{table}%
\begin{figure*}
        \centering
        \begin{subfigure}[b]{0.24\textwidth}
                \centering
                \includegraphics[width=\textwidth]{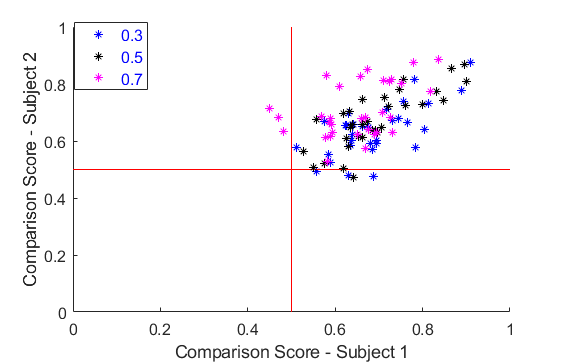}
                \caption{}
                \label{fig:gull}
        \end{subfigure}%
        \begin{subfigure}[b]{0.24\textwidth}
                \centering
                \includegraphics[width=\textwidth]{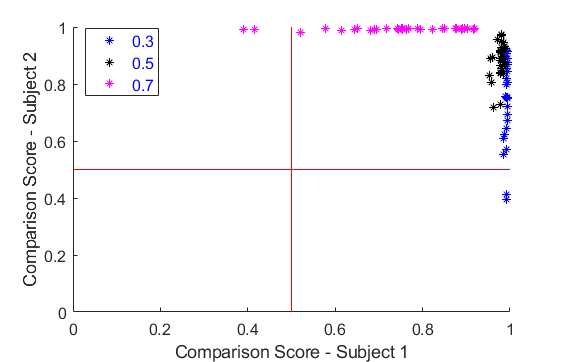}
                \caption{}
                \label{fig:gull2}
        \end{subfigure}%
        ~ 
        \begin{subfigure}[b]{0.24\textwidth}
                \centering
                \includegraphics[width=\textwidth]{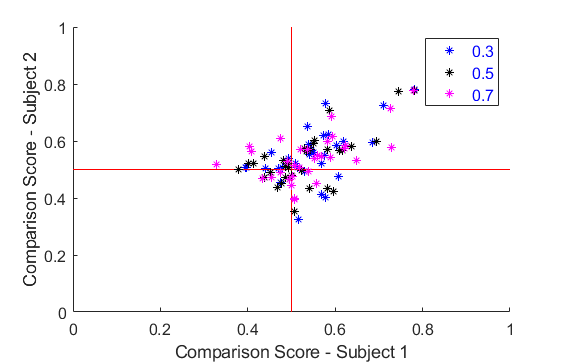}
                \caption{}
                \label{fig:tiger}
        \end{subfigure}
        ~ 
        \begin{subfigure}[b]{0.24\textwidth}
                \centering
                \includegraphics[width=\textwidth]{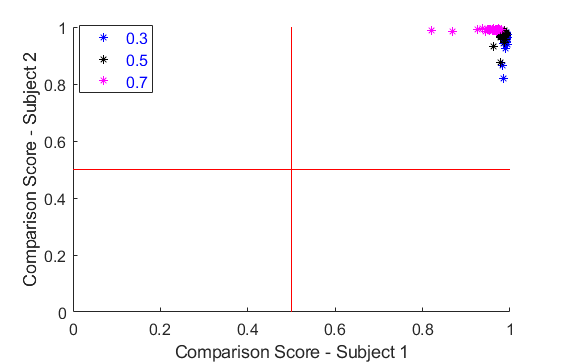}
                \caption{}
                \label{fig:mouse}
        \end{subfigure}
        \caption{Scatter plot of vulnerability when morphing images are from lookalike and identical twins (a) Lookalike: Arcface FRS (b) Lookalike: COTS FRS (c) Identical twins: Arcface FRS (d) Identical twins: COTS FRS }\label{fig:Vul1}
\end{figure*}
\begin{figure*}
        \centering
        \begin{subfigure}[b]{0.25\textwidth}
                \centering
                \includegraphics[width=\textwidth]{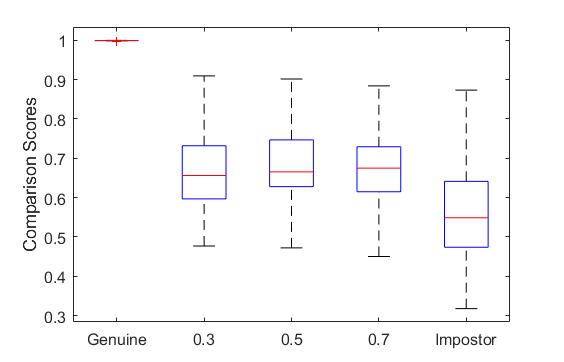}
                \caption{}
                \label{fig:gull}
        \end{subfigure}%
        \begin{subfigure}[b]{0.25\textwidth}
                \centering
                \includegraphics[width=\textwidth]{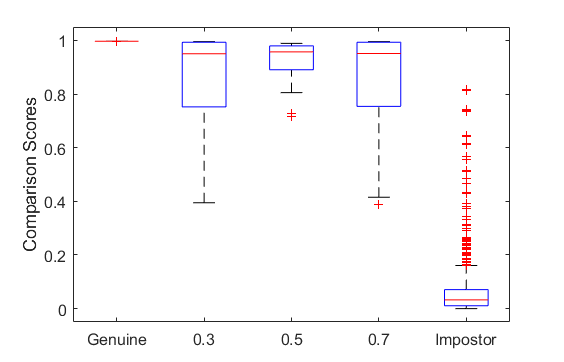}
                \caption{}
                \label{fig:gull2}
        \end{subfigure}%
          \begin{subfigure}[b]{0.25\textwidth}
                \centering
                \includegraphics[width=\textwidth]{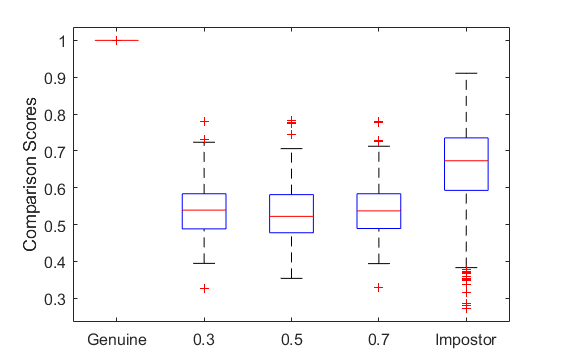}
                \caption{}
                \label{fig:gull2}
        \end{subfigure}%
           \begin{subfigure}[b]{0.25\textwidth}
                \centering
                \includegraphics[width=\textwidth]{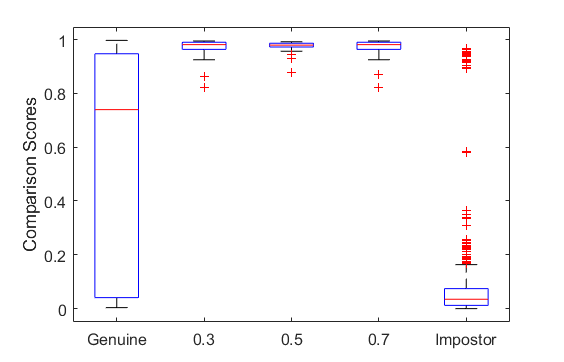}
                \caption{}
                \label{fig:gull2}
        \end{subfigure}%
               \caption{Box plot with verification scores   (a) Lookalike: Arcface FRS (b) Lookalike: COTS FRS (c) Identical twins: Arcface FRS (d) Identical twins: COTS FRS}\label{fig:BoxLookAlike}
\end{figure*}

\section{Experiments and Results}
\label{sec:exp}
Vulnerability analyses of identical twins and lookalikes are benchmarked and discussed in this section. To compute the vulnerability, we employed both commercial and deep-learning-based face recognition systems. The COTS system corresponds to Cognitec Face VACS SDK version 9.4.2 \cite{cognitec-FRS-SDK} and the deep-learning FRS is Arcface \cite{Deng-ArcFace-IEEE-CVPR-2019}. These two FRS are considered owing to their robustness and accurate face verification performance and are also  widely employed FRS for benchmarking the vulnerability  of face morphing techniques.

To quantitatively compute the vulnerability, we employed the generalized morphing attack potential (G-MAP) \cite{GMAP} which can quantify the vulnerability with a variable number of attempts against a given morphing image and across the different FRS while accounting for Failure To Acquire Rate (FTAR) and different types of morphing generation. In this work, we present the vulnerability results in two steps: (1) G-MAP with multiple attempts, in which the vulnerability is presented individually on each FRS for multiple attempts; (2)  G-MAP  with FTAR = 0 (as the FRS employed in this work can extract facial templates for all probe images) and a number of morphing types to one (as we have used only LMA-based morphing generation). For more information on G-MAP, refer to \cite{GMAP}.

In this paper, we present a vulnerability analysis of four different case studies. \textbf{Case-I:} The vulnerability of the lookalike faces is presented when morphing is generated between the lookalike pairs. \textbf{Case-II:}  The vulnerability of the non-lookalike faces is presented by generating the morphing faces from the non-lookalike data subjects from the same dataset. \textbf{Case-III:} The vulnerability of the identical twins is presented by generating the morphing images from the identical twins’ pairs. \textbf{Case-IV:} The vulnerability is computed on the non-identical twins morphing. These four case studies were designed to effectively benchmark the vulnerability of identical twins versus lookalike versus normal data. Furthermore, the analysis with three different morphing factors, 0.3, 0.5, and 0.7, is presented for all four case studies.

\begin{figure*}
        \centering
        \begin{subfigure}[b]{0.24\textwidth}
                \centering
                \includegraphics[width=\textwidth]{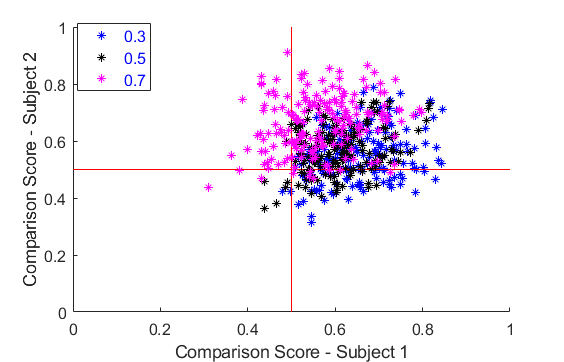}
                \caption{}
                \label{fig:gull}
        \end{subfigure}%
        \begin{subfigure}[b]{0.24\textwidth}
                \centering
                \includegraphics[width=\textwidth]{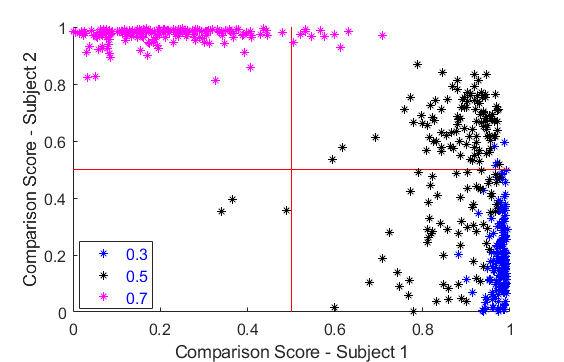}
                \caption{}
                \label{fig:gull2}
        \end{subfigure}%
        ~ 
        \begin{subfigure}[b]{0.24\textwidth}
                \centering
                \includegraphics[width=\textwidth]{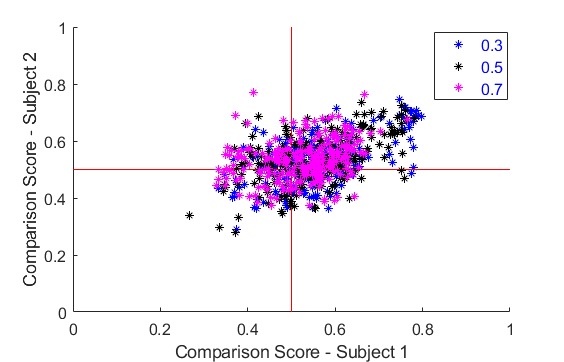}
                \caption{}
                \label{fig:tiger}
        \end{subfigure}
        ~ 
        \begin{subfigure}[b]{0.24\textwidth}
                \centering
                \includegraphics[width=\textwidth]{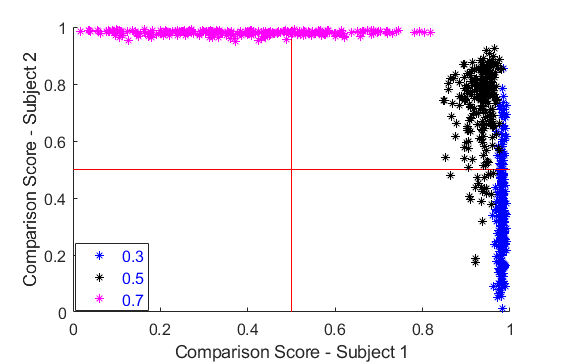}
                \caption{}
                \label{fig:mouse}
        \end{subfigure}
        \caption{Scatter plot of vulnerability when morphing images are from normal faces (a) Not Lookalike: Arcface FRS (b) Not Lookalike: COTS FRS (c) Not Identical twins: Arcface FRS (d) Not Identical twins: COTS FRS}\label{fig:Vul2}
\end{figure*}

\begin{figure*}
        \centering
        \begin{subfigure}[b]{0.25\textwidth}
                \centering
                \includegraphics[width=\textwidth]{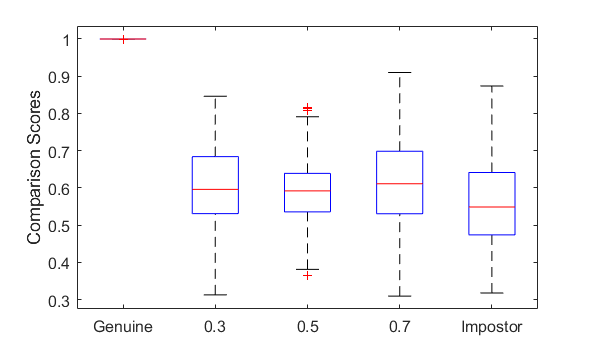}
                \caption{}
                \label{fig:gull}
        \end{subfigure}%
        \begin{subfigure}[b]{0.25\textwidth}
                \centering
                \includegraphics[width=\textwidth]{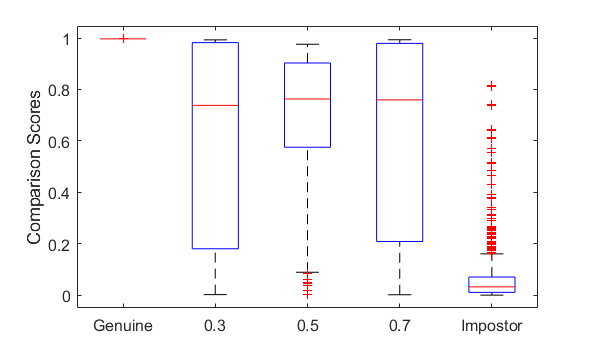}
                \caption{}
                \label{fig:gull2}
        \end{subfigure}%
         \begin{subfigure}[b]{0.25\textwidth}
                \centering
                \includegraphics[width=\textwidth]{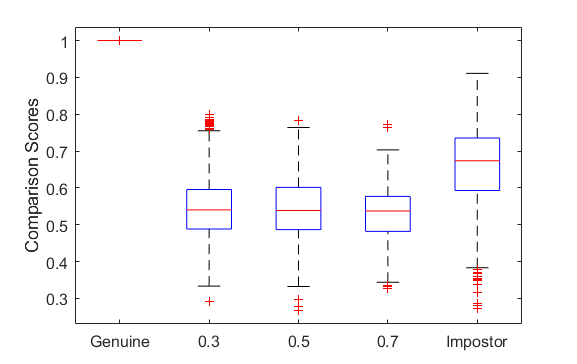}
                \caption{}
                \label{fig:gull2}
        \end{subfigure}%
             \begin{subfigure}[b]{0.25\textwidth}
                \centering
                \includegraphics[width=\textwidth]{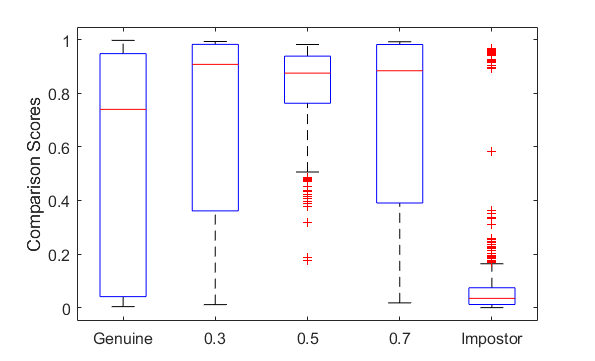}
                \caption{}
                \label{fig:gull2}
        \end{subfigure}%
               \caption{Box plot with verification scores (a) Not Lookalike: Arcface FRS (b) Not Lookalike: COTS FRS (c) Not Identical twins: Arcface FRS (d) Not Identical twins: COTS FRS}\label{fig:BoxTwins}
\end{figure*}
Figure \ref{fig:Vul1} shows the scatter plot of the comparison scores computed using two different FRS when enrolled with morphing images generated from lookalike twins and identical twins. Scatter plots are shown for different morphing factors such as 0.3, 0.5, and 0.7. The red lines in the scatter plots indicate the thresholds set at FAR = 0.1\%. Figure \ref{fig:BoxLookAlike} shows box plots of the verification scores computed using two different FRS and the morphing comparison scores corresponding to the three different morphing factors. Table \ref{fig:BoxTwins} lists the quantitative values of vulnerabilities computed using G-MAP (with multiple attempts). The following are the important observations:
\begin{itemize}
    \item It is interesting to observe that both lookalike and identical twins indicate vulnerability on FRS irrespective of the morphing factor. This can be observed in Figure \ref{fig:BoxTwins}, where the morphing scores corresponding to different morphing factors show similar distributions. The quantitative value of the vulnerability computed using G-MAP indicated in Table \ref{fig:BoxTwins} (refer to Case-I and III) also reflects less variation between the different morphing factors.
\item Both FRS systems indicate the vulnerability to lookalike and identical twins irrespective of the morphing factor. COTS indicates a higher vulnerability compared to Arcface FRS on both lookalike and identical twin morphing images.
\item COTS FRS indicates a higher vulnerability with identical twins, and Arcface FRS indicates a higher vulnerability with lookalike morphing. 
\item Among three different morphing factors, 0.5 indicates the highest vulnerability of FRS, followed by 0.3 and 0.7. The FRS indicates the 
lower vulnerability on lookalike faces than on identical twins,  especially with a morphing factor of 0.7.
 \item Figure \ref{fig:BoxTwins} shows the box plots of the genuine, impostor, and morphing scores from identical twins and lookalikes computed using COTS and Arcface FRS. The Arcface FRS indicated good verification performance on identical twins and lookalikes compared to the COTS FRS. The verification performance of the COTS FRS on identical twins shows degraded performance owing to the high overlap of the genuine and impostor scores.
\item The distribution of morphing scores (with different morphing factors) indicates the high overlapping with genuine scores, mainly with COTS FRS on identical twins and lookalikes. Therefore, COTS FRS is more vulnerable than Arcface, which is also quantitatively acknowledged in Table \ref{tab:FMMPMR} (refer to Cases I and III).  
\end{itemize}

Figure \ref{fig:Vul2} and \ref{fig:BoxTwins}  (Case-II and Case-IV) shows the scatter and box plot computed when morphing scores are computed on the morphing images are generated using normal (not lookalike and not identical twins) datasets. The critical observations are as follows.
\begin{itemize}
\item Morphing factor plays a vital role in achieving the vulnerability of FRS. A morphing factor of 0.5 indicates a higher vulnerability with both FRS.
\item Arcface FRS indicates a higher vulnerability than COTS FRS on Case-II. However, the COTS FRS indicated a higher vulnerability than the Arcface FRS with Case-IV.
\end{itemize}

\begin{table}[htp]
  \centering
  \caption{Quantitative Performance of the Vulnerability using G-MAP (\%) with multiple attempts}
    \resizebox{0.8\linewidth}{!}{
    \begin{tabular}{|c|c|c|c|}
    \hline
    \multicolumn{1}{|c|}{\multirow{2}[4]{*}{Case Study}} & \multicolumn{1}{c|}{\multirow{2}[4]{*}{Morphing Factor}} & \multicolumn{2}{p{8.43em}|}{G-MAP (\%) with multiple attempts} \bigstrut\\
\cline{3-4}          &       & \multicolumn{1}{p{4.215em}|}{ArcFace} & \multicolumn{1}{p{4.215em}|}{COTS} \bigstrut\\
    \hline
    \multicolumn{1}{|c|}{\multirow{3}[6]{*}{Case-I}} & 0.3   & 90.62 & 93.1 \bigstrut\\
\cline{2-4}          & 0.5   & 96.88 & 100 \bigstrut\\
\cline{2-4}          & 0.7   & 91.62 & 93.75 \bigstrut\\
    \hline
    \multicolumn{1}{|c|}{\multirow{3}[6]{*}{Case-II}} & 0.3   & 71.1  & 2.31 \bigstrut\\
\cline{2-4}          & 0.5   & 80.23 & 60.47 \bigstrut\\
\cline{2-4}          & 0.7   & 73.91 & 6.32 \bigstrut\\
    \hline
    \multicolumn{1}{|c|}{\multirow{3}[6]{*}{Case-III}} & 0.3   & 53.12 & 100 \bigstrut\\
\cline{2-4}          & 0.5   & 70.62 & 100 \bigstrut\\
\cline{2-4}          & 0.7   & 53.12 & 100 \bigstrut\\
    \hline
    \multicolumn{1}{|c|}{\multirow{3}[6]{*}{Case-IV}} & 0.3   & 52.55 & 21.96 \bigstrut\\
\cline{2-4}          & 0.5   & 60.39 & 91.8 \bigstrut\\
\cline{2-4}          & 0.7   & 49.61 & 28.91 \bigstrut\\
    \hline
    \end{tabular}%
    }
  \label{tab:FMMPMR}%
\end{table}%
Based on the extensive experiments reported in Table \ref{tab:FMMPMR} (also in Figures \ref{fig:Vul1},\ref{fig:BoxLookAlike},\ref{fig:Vul2} and \ref{fig:BoxTwins} it can be noticed that: 
\begin{itemize}
    \item FRS are highly vulnerable to the morphing images generated using lookalike and identical twins than normal faces. 
    \item Vulnerability of the lookalike and identical twins morphing are less influenced by the morphing factor than normal morphing. 
    \item COTS FRS is highly vulnerable on lookalike and identical twins compared to normal face morphing. 
\end{itemize}

\begin{table}[htbp]
  \centering
  \caption{Quantitative Performance of the Vulnerability using G-MAP(\%)}
  \resizebox{0.75\linewidth}{!}{
    \begin{tabular}{|c|r|r|}
    \hline
    \multicolumn{1}{|r|}{\multirow{2}[2]{*}{Case Study}} & \multicolumn{1}{r|}{\multirow{2}[2]{*}{Morphing Factor}} & \multicolumn{1}{r|}{\multirow{2}[2]{*}{G-MAP (\%)}} \bigstrut[t]\\
          &       &  \bigstrut[b]\\
    \hline
    \multicolumn{1}{|c|}{\multirow{3}[6]{*}{Case-I}} & 0.3   & 84.38 \bigstrut\\
\cline{2-3}          & 0.5   & 96.88 \bigstrut\\
\cline{2-3}          & 0.7   & 84.13 \bigstrut\\
    \hline\hline
    \multicolumn{1}{|c|}{\multirow{3}[6]{*}{Case-II}} & 0.3   & 2.31 \bigstrut\\
\cline{2-3}          & 0.5   & 48.26 \bigstrut\\
\cline{2-3}          & 0.7   & 4.89 \bigstrut\\
    \hline\hline
    \multicolumn{1}{|c|}{\multirow{3}[6]{*}{Case-III}} & 0.3   & 53.12 \bigstrut\\
\cline{2-3}          & 0.5   & 40.62 \bigstrut\\
\cline{2-3}          & 0.7   & 53.12 \bigstrut\\
    \hline\hline
    \multicolumn{1}{|c|}{\multirow{3}[6]{*}{Case-IV}} & 0.3   & 11.76 \bigstrut\\
\cline{2-3}          & 0.5   & 45.7 \bigstrut\\
\cline{2-3}          & 0.7   & 16.41 \bigstrut\\
    \hline
    \end{tabular}%
       }
  \label{tab:MAPl}%
\end{table}%
\begin{figure}[htp]
\begin{center}
\includegraphics[width=1.0\linewidth]{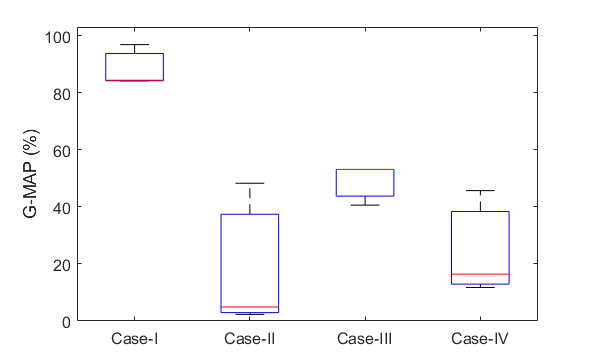}
\end{center}
   \caption{Box plot illustrating the vulnerability with G-MAP (\%)  with FTAR = 0 and Number of morphing types = 1 }
\label{fig:MAPBOX}
\end{figure}

 Table \ref{tab:MAPl} lists the quantitative values of G-MAP that provide an attack assessment across different FRS. 
 Figure \ref{fig:MAPBOX} shows box plots of the vulnerability with G-MAP for the four cases.  Based on the obtained results, the following can be observed.
\begin{itemize}
    \item The vulnerability of the FRS is high with a lookalike and identical twins morphing samples than normal (not lookalike and not identical twins) morphing samples.
\item Highest vulnerability of FRS is noted with the lookalike faces compared to the identical twins morphing images.
\end{itemize}

\subsection{Discussion}

Based on the observations from the experiments and obtained results, the research questions formulated in Section \ref{sec:intro} are answered below.
\begin{itemize}[leftmargin=*,noitemsep, topsep=0pt,parsep=0pt,partopsep=0pt]
    \item {\textbf{Q1}. Does the morphing of lookalike, and identical twins indicate higher vulnerability of FRS compared to the normal (or regular) face?}
    \begin{itemize}[leftmargin=*,noitemsep, topsep=0pt,parsep=0pt,partopsep=0pt]
        \item Based on the quantitative performance reported in Table \ref{tab:FMMPMR} indicates a higher vulnerability on lookalike and identical twins compared to the regular faces on both FRS employed in this work. Thus, our experimental results showed the higher vulnerability of FRS with lookalike and identical twins.   
    \end{itemize}
    \item {\textbf{Q2}. Does the morphing of lookalike data subjects indicate higher vulnerability of FRS  than identical twins?} 
        \begin{itemize}[leftmargin=*,noitemsep, topsep=0pt,parsep=0pt,partopsep=0pt]
            \item Based on the experimental results reported in Table \ref{tab:FMMPMR} and \ref{tab:MAPl}, the morphing of lookalike data subjects indicates higher vulnerability compared to the morphing of identical twins. 
        \end{itemize}
    \item {\textbf{Q3}. Does the morphing factor influence the vulnerability of FRS to lookalikes than identical twins?} 
        \begin{itemize}[leftmargin=*,noitemsep, topsep=0pt,parsep=0pt,partopsep=0pt]
            \item   Based on the extensive experimental results reported in Figure \ref{fig:Vul1}, \ref{fig:BoxLookAlike} and Table \ref{tab:FMMPMR}, morphing of lookalike and identical twins can indicate the vulnerability on the FRS equally with different FRS. A similar observation is not noticed with the normal (or regular) face morphing (based on Figure \ref{fig:Vul2} and \ref{fig:BoxTwins}).   
 \end{itemize}           
\item {\textbf{Q4}. Does the Commercial-Off-The-Shelf (COTS) FRS indicate a higher vulnerability than deep learning FRS (Arcface)?} 
        \begin{itemize}[leftmargin=*,noitemsep, topsep=0pt,parsep=0pt,partopsep=0pt]
            \item Based on the results reported in Table \ref{tab:FMMPMR} indicates the higher vulnerability of COTS than Arcface. 
        \end{itemize}
\end{itemize}
\section{Conclusions}
\label{sec:conc}
Evolving attacks on face recognition systems is a growing concern for achieving reliable and secure access control. The morphing attacks demonstrated the high vulnerability of the FRS and the human observers. In this work, we presented the first study on the morphing of real lookalikes and identical twins and their impact on the FRS. We introduced a new dataset constructed using lookalike and identical twin morphologies. The newly constructed datasets comprised normal (or regular) face morphing to effectively benchmark the vulnerability. Morphing was carried out using landmark methods with three different morphing factors: 0.3, 0.5, and 0.7. Extensive experiments were carried out using four different case studies (or evaluation protocols), indicating a higher vulnerability of FRS to lookalike and identical twins than normal morphing. Further analysis also indicated that the lookalike is more vulnerable than identical twin morphing. Future work will address the current limitations identified in this work, such as (a) increasing the size of the dataset, (2) extending the lookalike and identical twin morphing on print scan scenarios,  and (3) benchmarking morphing attack detection techniques.

{\small
\bibliographystyle{IEEEtran}
\bibliography{Face_Morph_references}
}
\end{document}